# AutoKE: An automatic knowledge embedding framework for scientific machine learning

Mengge Du, Yuntian Chen, and Dongxiao Zhang

*Abstract*—Imposing physical constraints on neural networks as a method of knowledge embedding has achieved great progress in solving physical problems described by governing equations. However, for many engineering problems, governing equations often have complex forms, including complex partial derivatives or stochastic physical fields, which results in significant inconveniences from the perspective of implementation. In this paper, a scientific machine learning framework, called AutoKE, is proposed, and a reservoir flow problem is taken as an instance to demonstrate that this framework can effectively automate the process of embedding physical knowledge. In AutoKE, an emulator comprised of deep neural networks (DNNs) is built for predicting the physical variables of interest. An arbitrarily complex equation can be parsed and automatically converted into a computational graph through the equation parser module, and the fitness of the emulator to the governing equation is evaluated via automatic differentiation. Furthermore, the fixed weights in the loss function are substituted with adaptive weights by incorporating the Lagrangian dual method. Neural architecture search (NAS) is also introduced into the AutoKE to select an optimal network architecture of the emulator according to the specific problem. Finally, we apply transfer learning to enhance the scalability of the emulator. In experiments, the framework is verified by a series of physical problems in which it can automatically embed physical knowledge into an emulator without heavy hand-coding. The results demonstrate that the emulator can not only make accurate predictions, but also be applied to similar problems with high efficiency via transfer learning.

*Impact Statement* — Embedding physical knowledge into machine learning has been widely applied in solving scientific computing problems. However, it is tedious and time-consuming to establish the emulator and embed physical knowledge into it. To this end, we proposed a plug-and-play framework for solving complicated physical problems. The user inputs equations with arbitrary form according to the specified rules, which can be automatically parsed and transformed into a structural representation that the program can recognize. The embedding of physical knowledge, the establishment of emulators, and the adjustment of parameters are automated, which provides great convenience while obtaining accurate solutions. Moreover, the strong scalability enables the trained emulator to be applied to similar problems, which dramatically improves efficiency. Overall, the proposed framework is both user-friendly and lowers the application threshold of related research.

*Index Terms*— knowledge embedding, physics informed neural network, automatic machine learning, equation parser, adaptive optimization parameters, transfer learning.

## I. INTRODUCTION

WITH the development of machine learning, and especially the progress and maturity of deep learning, many scholars have gradually utilized deep neural networks (DNNs) combined with domain knowledge to solve scientific computing problems [1], [2]. On the one hand, according to the universal approximation theorem [3], the powerful approximation capabilities of the neural network enable it to represent a wide range of functions and solve complex physical problems well. On the other hand, the embedding of physical knowledge, such as governing equations, can provide abundant prior information, accelerate training of the network, and prevent the model from overfitting [4]-[6]. Therefore, determination of how to effectively and efficiently embed domain knowledge into the neural networks is crucial for training a satisfactory model with common physical sense.

For solving PDEs or PDE-based physical problems, one of the effective and straightforward knowledge embedding methods is to constrain the neural network by incorporating the residual of PDEs (i.e., governing equations) and other boundary conditions into the loss function. Based on this idea, the physics informed neural network (PINN) was proposed [7], and its variants made further improvements [8]-[12] and have been extended to other equations, such as fractional differential equations [13]. For some actual engineering problems, such as subsurface flow in hydrology, Wang *et al.* put forward a theory-guided neural network (TgNN), which also integrates engineering control and expert knowledge into the modeling process [14]. In the implementation

This work is supported by National Center for Applied Mathematics Shenzhen （NCAMS), the Shenzhen Key Laboratory of Natural Gas Hydrates (Grant No. ZDSYS20200421111201738), and the SUSTech - Qingdao New Energy Technology Research Institute. *(Corresponding author: Dongxiao Zhang.)*

M. Du is with the College of Engineering, Peking University, Beijing 100871, China.

Y. Chen is with the Eastern Institute for Advanced Study, Yongriver Institute of Technology, Zhejiang 315201, China.

D. Zhang is with the National Center for Applied Mathematics Shenzhen (NCAMS), Southern University of Science and Technology, Guangdong 518055, China and the Department of Mathematics and Theories, Peng Cheng Laboratory, Guangdong, China (e-mail: zhangdx@sustech.edu.cn).

stage, each physical constraint is represented as a loss term, and each loss item is assigned a certain weight in the loss function. The prediction that violates the physical constraints is gradually corrected by the gradient-based optimization method. It can be seen that implementation of these soft constraints is easy, and these constraints are proven to be effective if appropriate weights are utilized. The rapid development of this knowledge embedding method is largely due to the progress of related frameworks, especially the development of automatic differentiation (AD) technology [15], which greatly simplifies calculation of the PDE residual.

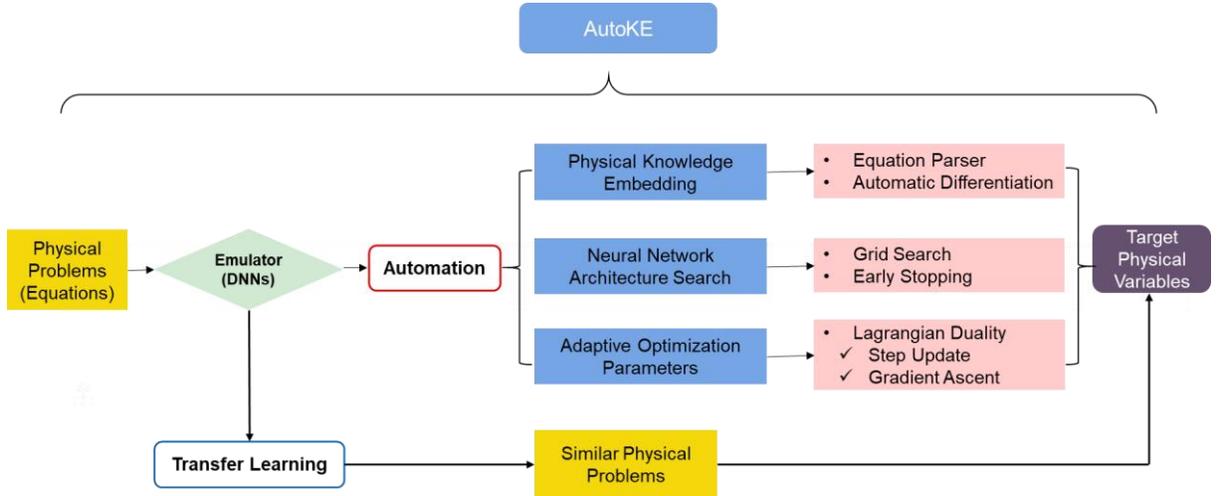

Fig. 1. Schematic of the AutoKE framework.

Nevertheless, it remains a major challenge to build a suitable neural network-based emulator for some physical problems with complex equation forms. First, researchers who intend to use deep learning to solve engineering problems not only need to possess a certain foundation of physical knowledge in this field, but are also required to master theoretical knowledge related to neural networks. Moreover, they are also required to master the use of deep learning frameworks, such as PyTorch [16] and TensorFlow [17], which have a steep learning curve and application difficulty. Although some PINN-related frameworks have been proposed, such as DeepXDE [18], they are often limited to well-defined physical problems and are not well-solved for problems constrained by complex equations. Specifically, although the derivation of partial differential terms can be obtained manually and the implementation of physical constraints is relatively straightforward, users are often forced to complete the entire modeling and coding process by themselves, which is tedious and inconvenient.

Furthermore, there are still some unresolved problems in existing methods: (1) weights of different loss terms in the loss function generally need to be manually adjusted by researchers based on experience. It is necessary to conduct repetitive experiments, which is very time-consuming. (2) Determination of the neural architecture is relatively casual and often has not been carefully adjusted. For problems of different complexities, the expressivity requirements of the network may be different, and the training error and accuracy of the prediction results may be affected accordingly. In addition, physical problems in the same field often have similar forms of governing equations, except that some equation terms are added or some physical coefficients are changed. However, training a large number of neural networks to solve similar problems is expensive and inefficient. Some studies have realized the importance of transfer learning [19], [20] and started to use it to speed up training in different physics problems [21], but related research is still relatively unfocused and unsystematic.

To solve these problems, this paper proposes an automatic and efficient scientific machine learning framework that can automatically implement physical knowledge embedding for complex physical problems containing complex compound partial derivatives or stochastic physical fields. Moreover, based on the Lagrangian dual method, adaptive optimization parameters are introduced, which enables the weights to be adjusted dynamically according to their corresponding loss without manual intervention. In addition, the grid search algorithm and early stopping are utilized to find architectures that achieve satisfying performance on specific physical problems within a short time period. By using this framework, accurate prediction results can be obtained, and at the same time transfer learning [20], [22] is utilized to increase the scalability of the trained model and improve training efficiency for solving similar problems. We also propose a new approach to complex problems using transfer learning from the perspective of curriculum learning [23]. Finally, several physical problems, including the Korteweg-de Vries equation (KdV), viscous gravity current, and subsurface reservoir flow, are utilized to test the performance of the framework.

The remainder of this paper proceeds as follows. In section II, we first review the process of solving physical problems based on PINN and introduce our motivation for building a new framework - AutoKE - by taking subsurface reservoir flow as an example. Then, the main components and algorithms of the AutoKE framework are introduced in section III, and the specific application process is explained. The performance of the framework on several physical equations and engineering problems is presented in section IV, which highlights the ease of use and prediction accuracy of the framework. Finally, we conclude our paper and introduce our future work in section V.

## II. PROBLEM ANALYSIS AND MOTIVATIONS

In this section, we first review the framework of PINN and its physical knowledge embedding method, and then introduce the subsurface reservoir flow problem, which is used as an example to explain that the existing framework is neither sufficient nor perfect. Finally, the motivation of the AutoKE framework is presented.

### A. PINN

Compared with the common data-driven approach, PINN imposes physical constraints or knowledge on the neural network by utilizing governing equations. Consider a problem or PDE as follows:

$$\begin{cases} \mathrm{F}\,(u(\mathbf{x},t)) = 0, & \Omega \times [0,T] \\ \mathrm{B}(u(\mathbf{x},t)) = 0, & \partial\Omega \times [0,T] \\ u(\mathbf{x},0) = u_0(\mathbf{x}), & \Omega \times \{t=0\} \end{cases} \tag{1}$$

where $\mathrm{F}$ denotes the nonlinear differential operator; $u(x,t)$ denotes the state of interest; $\mathrm{B}$ denotes the boundary operator; $\Omega \subset \Box^d$ denotes the space domain; and $T$ denotes the time domain.

Deep neural networks (DNNs) are often utilized to approximate $u(\mathbf{x},t)$. The input of the DNN is the space and time coordinates represented as a vector $\mathbf{x}$ and $t$, respectively. The output is the state of interest $u$. A DNN generally has several layers and incorporates a nonlinear activation function $\sigma$ to strengthen its fitting ability. A DNN used in PINN can be constructed as follows:

$$\mathbf{u} = \sigma(\mathbf{W_{L\text{-}1}}\sigma(\mathbf{W_{L\text{-}1}}(\dots\,\sigma(\mathbf{W_i}\mathbf{x}+\mathbf{b_1}))+\mathbf{b_{L\text{-}1}})+\mathbf{b_L}) \tag{2}$$

where $\mathbf{W_i}$ and $\mathbf{b_i}$ denote the weights and bias of the $i$th layer of the feedforward neural network, respectively. The network parameters can be represented as $\theta = \{\mathbf{W_i}, \mathbf{b_i}\}_{i=1}^{L+1}$.

In order to learn a suitable set of parameters for the neural network to predict the solution, a loss function combined with physical information needs to be formulated. Based on the mean squared error (MSE), it can usually be defined as follows:

$$\mathrm{L}(\theta) = \lambda_r \mathrm{L}_r + \lambda_{IC} \mathrm{L}_{IC} + \lambda_{BC} \mathrm{L}_{BC} + \lambda_{Data} \mathrm{L}_{Data} \tag{3}$$

where $\mathrm{L}_r$ denotes the loss term that measures the residual of the PDE system; $\mathrm{L}_{BC}$, $\mathrm{L}_{IC}$, and $\mathrm{L}_{Data}$ measure the MSE of boundary condition, initial condition, and labeled data, respectively; and $\lambda_r$, $\lambda_{BC}$, $\lambda_{IC}$, and $\lambda_{Data}$ denote the hyperparameters that measure the importance of the corresponding loss term to the final loss.

The trainable parameters $\theta$ can be adjusted by minimizing the loss function through some gradient-based optimization methods. After finishing the training process, the neural network can be regarded as a solver to solve its corresponding equation within the simulation domain.

### B. Problem analysis

Although PINN and its variants have achieved great success in solving ordinary differential equations (ODEs) and even some PDEs, the whole process of building a solver is complicated and cumbersome, and certain unsolved problems remain. We take an engineering example of subsurface single-phase reservoir flow as an example for further explanation.

In this part, a dynamic subsurface time-dependent single-phase flow problem in heterogeneous media is considered and satisfies the following governing equation:

$$\nabla \cdot (f(u)K(\mathbf{x})\nabla u(\mathbf{x},t)) = u_t \tag{4}$$

subject to the Dirichlet and Neumann boundary conditions:

$$u(\mathbf{x},\mathrm{t}) = h_D(\mathbf{x}), \ \mathbf{x} \in \Gamma_D \tag{5}$$

$$\boldsymbol{n} \cdot K(\mathbf{x})\nabla u(\mathbf{x}) = g(\mathbf{x}), \ \mathbf{x} \in \Gamma_N \tag{6}$$

and initial condition:

$$u(\mathbf{x},0) = h_0(\mathbf{x}) \tag{7}$$

where $f(u)$ denotes a known state-dependent constitutional relationship; $K(\mathbf{x})$ denotes a space-dependent diffusion coefficient, which usually refers to hydraulic conductivity; $\boldsymbol{n}$ is the unit normal vector to the Neumann boundary $\Gamma_N$; $\Gamma_D$ is the Dirichlet boundary; and $h_0(\mathbf{x})$ is the initial condition.

To solve this problem, we intend to construct a DNN to approximate $u(\mathbf{x},t)$ in the whole simulation domain. However, there are some difficulties to use the PINN framework directly.

Firstly, as mentioned previously, it is necessary and beneficial to embed physical knowledge into the training of DNNs. The most direct method is to add the MSE of the residual as a penalty to the loss function. Specifically, it is common to use the AD in a deep learning framework, such as PyTorch, to obtain those partial derivatives in the equation. It is worth noting that in this problem, the partial differential term $\nabla \cdot (f(u)K(\mathbf{x})\nabla u(\mathbf{x},t))$ is complex in form, and especially $f(u)$ may involve complex constitutive relationships. When calculating residuals, hand-coded analytical derivatives are error-prone and tedious, especially for those beginners who are not familiar with deep learning frameworks and AD.

In addition, the heterogeneous diffusion coefficient $K(\mathbf{x})$ is a function of space. We may know the value $K$ at each coordinate point, but its partial derivative $K_{\mathbf{x}}$ is unknown. Consequently, the finite difference (FD) method [24] is a common method to help approximate it. The whole process is complicated and tedious, and it is not covered in the regular PINNs framework. However, some researchers have proposed that an auxiliary DNN can be used to fit the unknown physical relationship so that the residual can be calculated by AD [25].

Apart from the difficulties mentioned above, there are some inherent problems of PINNs: (1) the loss function contains multiple loss items, and the weight of each loss item often needs to be adjusted by experienced researchers for repetitive experiments to ensure the stability of training and the accuracy of predictions. Some researchers have considered abandoning the method of fixed weights. For instance, Braga *et al.* [26] introduced a self-adaptive parameter adjustment method, and each weight in the loss function is trainable, forcing the neural network to focus on those loss items with larger values during optimization. Some scholars have transformed the original problem into a Lagrangian duality problem [27], [28]. It is also expected that weights in the loss function can be dynamically adjusted to reduce manual intervention. (2) Another problem to which many studies have not paid sufficient attention is the structure of DNNs. Although the default structure used in PINNs is the fully connected neural network, which is not complicated, there are still several hyperparameters that need to be adjusted, including activation functions and hidden unit size. For different problems, it is necessary to design a neural network with sufficient expressivity. (3) At the same time, for similar problems, no in-depth analysis of the scalability of the trained model yet exists. How to avoid repetitive training and improve training efficiency are also urgent problems to be solved.

### C. Motivations of the AutoKE framework

Through the above analysis, we conclude that the existing PINN methods are not perfect for solving the single-phase reservoir flow problem. Although most of the PINN-related toolboxes provide abundant examples, they are basically designed for specific equations and fail to solve equations with arbitrary form (i.e., open-form PDEs [29]). More importantly, the threshold for application is high, and users are forced to accomplish the physical knowledge embedding themselves.

To address the problems raised above, it is necessary to propose a new machine learning framework for physical knowledge embedding, which can assist users to impose physical constraints on DNNs automatically. Specifically, according to the input equation, the framework can analyze the physical quantities and relationships between different terms in the equation, and then realize automatic physical knowledge embedding, i.e., calculating the equation residual. This process necessitates a combination of an equation parser and AD in PyTorch. For complex problems, such as single-phase reservoir flow, the process of establishing the emulator is simplified to the greatest extent. In addition, for problems involving unknown physical relationships, a series of auxiliary DNNs should be established to fit the corresponding physical parameters, such as $K$, and the original values are substituted with the results of the neural network output so that the residuals can be quickly calculated by AD. For other inherent problems of PINN, we consider solving them by introducing adaptive parameters, neural architecture search (NAS), and transfer learning, thereby establishing a user-friendly framework for solving physical problems efficiently and automatically.

## III. AutoKE Framework

In this section, we provide a thorough introduction of our proposed framework based on the single-phase reservoir flow problem introduced in section II.

### A. Overview of the AutoKE framework

As shown in Fig. 1, compared to the PINN, the main features of AutoKE are automation and efficiency. When a physical problem is encountered, which is usually represented as some governing equations and boundary conditions, we intend to build a DNNs-based emulator or solver to predict the state of interest in the simulation domain. During the modeling process, automation mainly includes the following three aspects:

- Automatic imposition of physical constraints is accomplished by employing the equation parser and AD. The computational graph is constructed automatically to calculate the residual of the governing equation.

- By converting the original optimization problem into a Lagrangian dual problem, the weight of each loss term in the loss function is trainable or adjustable, and two update strategies are proposed.

- Based on the idea of NAS [30], AutoKE adopts the grid search algorithm [31] to find the network structure suitable for the current problem and utilizes the early stopping strategy to save searching time.

Efficient training is reflected in the fact that the existing framework supports the transfer of emulators using the transfer learning method for similar problems, which can significantly speed up training and improve efficiency.

### B. Automatic physical knowledge embedding

The physical constraints imposed on the neural network are mainly reflected in incorporating the residuals of the governing equations into the loss function. Therefore, for complex equation forms, such as single-phase reservoir flow, determination of how to correctly and automatically solve the residuals becomes the key to knowledge embedding. The flowchart of imposing physical constraints is presented in Fig. 2.

#### 1) DNNs

Unlike the common PINN frameworks that only make use of the prediction network to approximate the target state of interest, AutoKE also designs a series of fitting networks to fit unknown physical relationships. It can be seen from part I in Fig. 2 that we have DNNs predict the $u(t, x)$, and other parameters $v_1, v_2, ..., v_m$ can also be modeled with auxiliary DNNs. From the above analysis of single-phase flow, the unknown physics refers to the stochastic permeability field, and we can incorporate a DNN to fit $K(\mathbf{x})$, which is a function of space. Therefore, we have the following:

$$u(\mathbf{x}, t) \approx u(\mathbf{x}, t, \theta), \quad K(\mathbf{x}) \approx K(\mathbf{x}, \gamma) \tag{8}$$

where $\theta$ and $\gamma$ denote the parameters of the prediction network and fitting network, respectively. The spatial derivatives of $K(x, \gamma)$ can be obtained by AD, and the residual of the governing equation is represented as:

$$\mathrm{R} = \nabla \cdot (f(u(\mathbf{x}, t, \theta)) K(\mathbf{x}, \gamma) \nabla u(\mathbf{x}, t, \theta)) - u_t(\mathbf{x}, t, \theta) \tag{9}$$

and then physics constraints loss can be incorporated as follows:

$$\mathrm{L}_r = \frac{1}{N_C} \sum_{i=1}^{N_C} \left| \mathrm{R}\left(\mathbf{x}_i^c, t_i^c, \theta, \gamma\right) \right|^2 \tag{10}$$

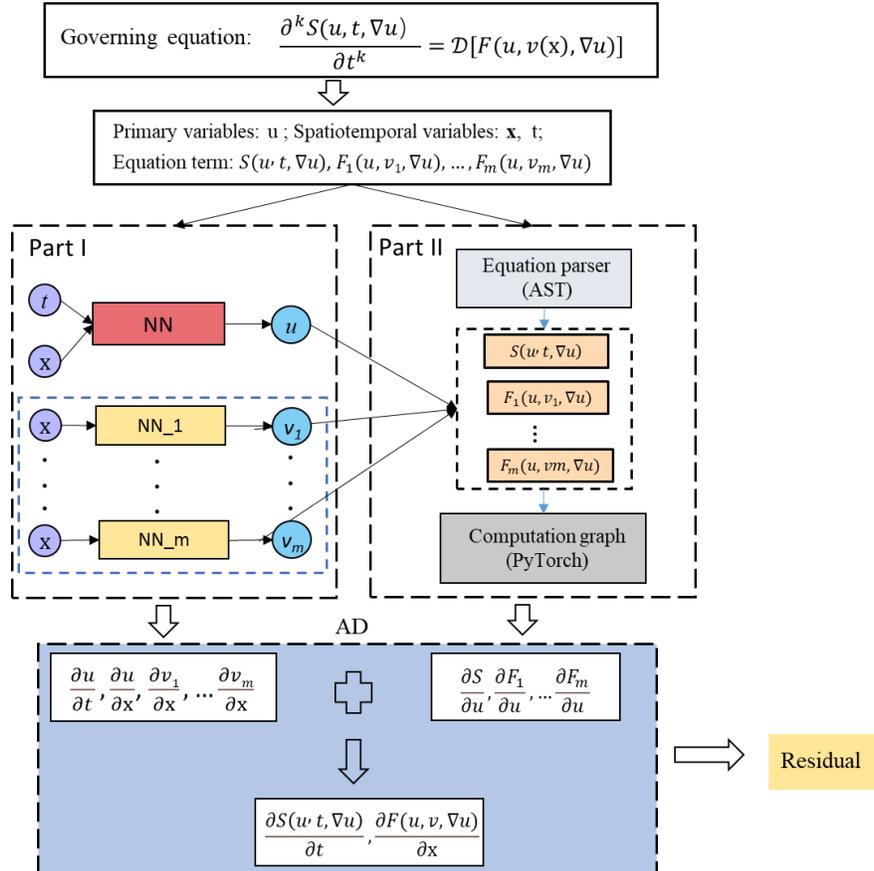

Fig. 2. Flowchart of automatically imposing physical constraints in the neural network.

where $\{\mathbf{x}_i^{\,c}, t_i^{\,c}\}_{i=1}^{N_c}$ is the collocation points; and $N_c$ denotes the total number of collocation points. The fitting network should also be forced to match the K measurements, and thus we have that $\{\mathbf{x}_i^{\,K}\}_{i=1}^{N_K}$ is the labeled training points and $\kappa^*$ denotes the $K$ measurement. If $u$ measurements are also available, we have:

$$L_{Data} = \frac{1}{N} \sum_{i=1}^{N} \left| u(\mathbf{x}_i^{\,N}, t_i^{\,N}, \theta) - u^*(\mathbf{x}_i^{\,N}, t_i^{\,N}) \right|^2 \tag{11}$$

where $\{\mathbf{x}_i^{\,N}, t_i^{\,N}\}_{i=1}^{N}$ denotes the labeled data of $u$ measurements.

According to the boundary conditions and initial conditions described in equation (4)-(7), their corresponding loss terms can also be expressed as follows:

$$L_{IC} = \frac{1}{N_{IC}} \sum_{i=1}^{N_K} \left| u(\mathbf{x}_i^{\,0}, 0, \theta) - h_0(\mathbf{x}^0) \right|^2 \tag{12}$$

$$L_{BC} = \frac{1}{N_D} \sum_{i=1}^{N_D} \left| u(\mathbf{x}^d, t^d, \theta) - h_D(\mathbf{x}^d) \right|^2 + \frac{1}{N_N} \sum_{i=1}^{N_N} \left| u(\mathbf{x}^n, t^n, \theta) - g(\mathbf{x}^n) \right|^2 \tag{13}$$

where $\{\mathbf{x}_i^{\,d}, t_i^{\,d}\}_{i=1}^{N_D}$ and $\{\mathbf{x}_i^{\,n}, t_i^{\,n}\}_{i=1}^{N_N}$ denote the collocation points at Dirichlet and Neumann boundary segments, respectively; and $\{\mathbf{x}_i^{\,0}, t_i^{\,0}\}_{i=1}^{N_{IC}}$ denotes the collocation points at the initial time. Then, the total loss function can be determined with the weighted sum of all of the loss terms above and written as:

$$L(\theta, \gamma) = \lambda_{data} L_{Data} + \lambda_K L_K + \lambda_r L_r + \lambda_{IC} L_{IC} + \lambda_{BC} L_{BC} \tag{14}$$

By minimizing the loss function above with a gradient-based optimization method, such as Adam [32], the prediction network and fitting network can be trained simultaneously.

### 2) Equation parser

To automate the process of calculating residuals, the equations should be parsed, and the information within the equations needs to be extracted. In the AutoKE framework, we define an equation parser module to accomplish this, which is illustrated in part II of Fig. 2. Users are required to input an ASCII representation of equations in a mathematical form. By predefining the primary variables of interest, and spatiotemporal variables, all of the components, including parameters and operators, are tractable, and an abstract syntax binary tree (AST) containing all of the fundamental information of the problem can be constructed. This tree representation makes it more flexible and convenient to deal with complex equations, and has been used in many other problems, such as PDE discovery [29], [33]. Then, post-order traversal of its nodes can generate the output code in PyTorch, which automatically builds a computational graph at the same time.

A top-down parser follows procedure 1 to parse the equation and generate the output binary tree. Fig. 3 presents an example of logical flow parsing one of the compositions in equation (4). The final abstract syntax tree generated for two-dimensional single-phase reservoir flow of equation (4) is shown in Fig. 4.

By combining part I and part II, a complete computational graph in PyTorch is constructed, and all of the required derivatives can be obtained through AD. It can be seen that, compared with manual handwritten code, the whole process of calculating the residual is greatly simplified and completely automatic.

| **Procedure 1**: Workflow of the equation parser |
| --- |
| **Step 1**    Define the composition of a specific equation (e.g., $f$ or $K$ in equation (4)), the primary variable $\mathbf{u}$, and the spatiotemporal variables $\mathbf{x}$ and $t$. |
| **Step 2**    Input an ASCII representation of the equation. |
| **Step 3**    Parse the equation and construct a binary abstract syntax tree representing the equation. |
| **Step 4**    Traverse through the abstract syntax tree using post-order traversal (e.g., traversal order of AST in Fig. 3: $K \rightarrow f \rightarrow * \rightarrow u \rightarrow \mathbf{x} \rightarrow diff \rightarrow * \rightarrow \mathbf{x} \rightarrow diff$) to build a computational graph by using operators of PyTorch. |

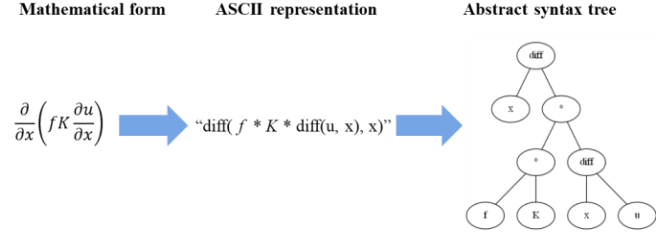

Fig. 3. An example of transforming an equation term into AST.

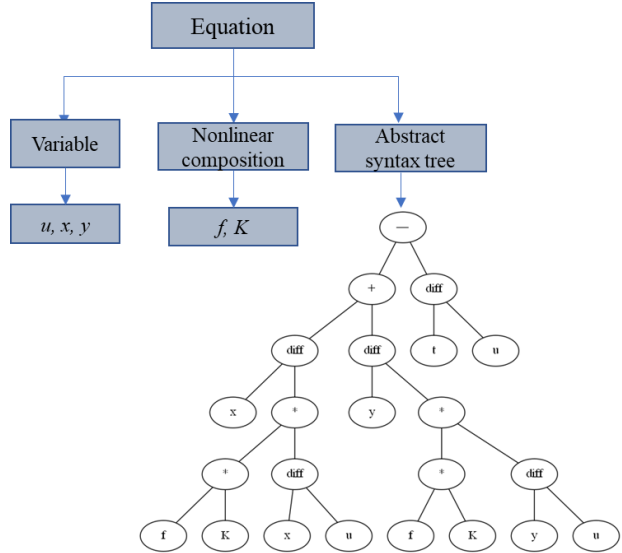

Fig. 4. An abstract syntax tree is created by the equation parser for two-dimensional single-phase reservoir flow.

### C. Neural architecture search

Deep learning has achieved remarkable success in the fields of images and natural language processing due to the delicate design of neural architecture [34]. Designing a suitable network structure is often accomplished by human experts and requires repeated experiments. The purpose of NAS is to automate the design process of the neural architecture [30], [34]. Although most of the neural networks used to solve physical problems in the PINNs framework are fully connected networks, it is still necessary to design a network with appropriate expressivity and structure according to the difficulty level of the problems. In the AutoKE framework, we make use of grid search algorithm and early stopping to help find the optimal hyperparameters. For example, common DNNs have three hyperparameters that need to be tuned: layer number of fully-connected neural network $L$; hidden units size $h$; and activation function type ($ACT$). Users are required to provide a specified finite subset of the hyperspace, for example:

$$L \in \{3, 4, 5\}$$
$$h \in \{32, 64, 128\}$$
$$ACT \in \{\text{Tanh}, \text{ReLU}, \text{GeLU}\}$$

Then, the framework performs an exhaustive search through the subset and trains DNNs with each pair $(L, h, ACT)$ selected from the three sets above. Each pair of parameters is also evaluated on a held-out validation set, and grid search outputs the combination of parameters with the best performance on the validation set at the end. If the search spaces of hyperparameters are sufficiently large, users can also adopt early stopping to save computational costs and accelerate the training process.

### D. Adaptive optimization parameters

In most cases, hard-coding weights are used in the loss function in equation (18) in the training process. Here, in the AutoKE framework, the weights are treated as hyper-parameters and are optimized during the training period. To accomplish this, we first convert the original problem into a constrained optimization problem by incorporating the Lagrangian dual method [27]. Since for most physical problems, including the single-phase reservoir flow problem, the weights of loss terms about initial conditions, boundary conditions, and data mismatch are less complicated with consistent physical unit and dimension, we keep $\lambda_{IC} = \lambda_{BC} = \lambda_{Data} = 1$ and only $\lambda_K$ and $\lambda_r$ are required to be optimized. The Lagrangian loss function can be rewritten as:

$$\min_{\theta, \gamma} L = L_{Data} + L_{IC} + L_{BC}$$

$$s.t. \begin{cases} L_K = 0 \\ L_r = 0 \end{cases}$$

(15)

By incorporating Lagrangian multipliers, it can be expressed as follows:

$$L_\lambda(\theta, \gamma) = L_{Data} + L_{IC} + L_{BC} + \lambda_K L_K + \lambda_r L_r$$

(16)

The loss (20) is first maximized with respect to the multipliers $\lambda_r$ and $\lambda_K$ and then minimized to update the trainable parameters $\theta$ and $\lambda$ of the neural network. This optimization process can be written as:

$$\min_{\theta, \gamma} \max_{\lambda_r, \lambda_K} L_\lambda(\theta, \lambda)$$

(17)

Instead of solving the primal above, we can convert it into a dual problem:

$$\max_{\lambda_r, \lambda_K} \min_{\theta, \gamma} L_\lambda(\theta, \lambda)$$

(18)

The multipliers should be non-negative and non-decreasing during the training process. We need to search a saddle point by means of repeated updates and, at each iteration, the multipliers will increment faster if their corresponding violation is larger. There are two ways to update the multipliers here. Taking the multiplier $\lambda_r$ as an example, one is updated according to a preset step size, which can be expressed as:

$$\lambda_r^{n+1} = \lambda_r^n + s_k \sum_{i=1}^N L_r(\mathbf{x}_i, t_i)$$

(19)

where $n$ refers to the current iteration; $N$ denotes the total number of data samples if the mini-batch technique is utilized; and $s_k$ denotes the step size.

The other is based on the idea of gradient ascent, which regards the multipliers as trainable parameters, and sets a new optimizer and learning rate in PyTorch to update it. The update is given by:

$$\lambda_r^{n+1} = \lambda_r^n + \alpha \nabla_{\lambda_r} L_r(\lambda_r, \lambda_k, \theta_k, \gamma_k)$$

(20)

where $\alpha$ denotes the learning rate at step $k$. It is worth noting that the direction of gradient update is opposite to the direction of parameters of the network update.

*E. Transfer learning*

As a method to speed up the convergence and improve the generalization performance of the model, transfer learning has achieved great success in the field of deep learning [19], [20], and especially in the field of computer vision [35]. Its general idea is to first train a fully convergent network on a basic task with a large amount of available data. The features of the trained network are then used as the initialization of the new network, and the training continues on the new task until convergence. Using transfer learning can often achieve better training and prediction results than random initialization. In the AutoKE framework, we provide interfaces for solving similar problems by using transfer learning methods to speed up training. Moreover, for the transferred features, the user is provided with a free choice, although in most cases, using all of the transferred features and fine-tuning the new task will achieve better results. In addition to improving the efficiency of training, we give another usage of transfer learning from the perspective of curriculum learning. When we solve a more complex physical problem, we can start with a simple basic problem. For difficult problems, this initialization method is often more conducive to the training of the network and avoids obtaining a locally optimal solution. Taking the single-phase reservoir flow problem as an example, in the experimental part, the transfer learning method is used to transfer the trained features to the model of the new task by taking a simple groundwater flow problem as the basic task, which improves the prediction performance and increases the speed of convergence significantly.

IV. NUMERICAL EXPERIMENTS

In this section, we evaluate the effectiveness and efficiency of the framework by conducting a series of numerical experiments on different physical problems, including the Korteweg-de Vries equation (KdV) and viscous gravity current. Furthermore, based on a two-dimensional subsurface single-phase reservoir flow, it is shown that Lagrangian dual optimization can realize adaptive weight adjustment, and NAS facilitates the process of finding the most suitable neural networks and significantly conserves training time. Finally, from the perspective of transfer learning, the extensibility and applicability of existing frameworks to well-trained emulators are demonstrated.

To evaluate the performance of the AutoKE framework, some metrics are introduced. The absolute error (AL) and $L_2$ relative error are expressed as:

$$AL(u_{pred}, u_{true}) = \left\| u_{pred} - u_{true} \right\|_2 \tag{21}$$

$$L_2(u_{pred}, u_{true}) = \frac{\left\| u_{pred} - u_{true} \right\|_2}{\left\| u_{true} \right\|_2} \tag{22}$$

where $\left\| \square \right\|$ denotes the standard Euclidean norm; $u_{pred}$ denotes the values predicted from the emulator; and $u_{true}$ denotes the reference values. The coefficient of determination, which is termed $R^2$ score, is also utilized as follows:

$$R^2 = 1 - \frac{\sum_{n=1}^{N_R} (u_{pred} - u_{true})^2}{\sum_{n=1}^{N_R} (u_{true,n} - \overline{u_{true}})^2} \tag{23}$$

where $N_R$ denotes the total number of points used for evaluation; and $\overline{u_{true}}$ denotes the average value of reference values.

It is worth noting that all experimental codes are based on the PyTorch environment, and both training and testing are performed on an NVIDIA GeForce RTX 1080 Ti.

### A. Korteweg-de Vries (KdV) equation

The one-dimensional Korteweg-de Vries (KdV) equation [36] is utilized as an instance to confirm that the AutoKE framework can not only automatically realize the establishment and solution of surrogate models, but can also handle nonlinear PDEs with high order derivatives. This equation models waves on shallow water surfaces and is related to several physical problems, including long internal waves and ion acoustic waves in plasma. Although the form of the equation is not complicated, there is a third-order derivative term in the equation, which increases the volatility and difficulty of solving the equation. The specific form of the KdV equation is as follows:

$$u_t = auu_x + bu_{xxx} \tag{24}$$

where $a$ and $b$ are constant parameters with $a = -1$ and $b = -0.0025$ in this example. The ASCII representation of the equation (24) residual is "a * u * diff(u,x) − b * diff(u,x,3) − diff(u,t)", which can be automatically transformed into a structural representation after being input to the equation parser, and it can be easily recognized by the program and used for constructing the computational graph in PyTorch.

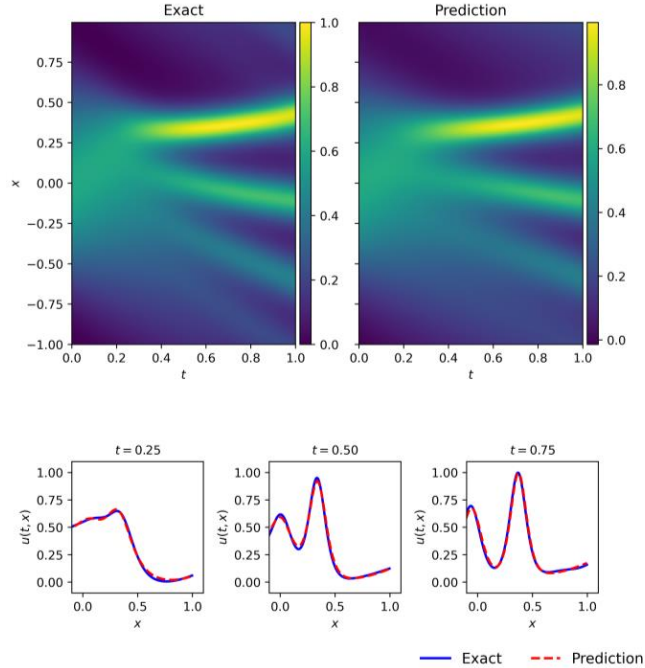

Fig. 5. Solutions of the KdV equation learned using the AutoKE framework vs. the exact solution.

To obtain the training and testing data, the spectral method is employed. In this problem, training data start from an initial condition $u(0, x) = \cos(\pi x)$, and are selected from 512 spatial observation steps and 201 temporal observation steps. We randomly sample 200 boundary points on the boundary $x = -1$ and $x = 1$, and 200 initial points and 10,000 collocation points on the rest of the simulation domain. The neural network implemented for this problem has 6 hidden layers with 128 neurons per layer, and Tanh is taken as the activation function. During the training process, the Adam optimizer [32] is first used to optimize the loss function, and the L-BFGS quasi-newton method [37] is then used to further fine-tune the neural network. Specifically, the weights in the loss function are set to 1, since it can obtain satisfactory results with the default settings.

As illustrated in Fig. 5, the top panel shows the reference solution of the equation and the result learned by the emulator. It can be seen that the results of the two are highly similar in the entire data domain with a relative $L_2$ error of 3.9E-2. Specifically, we also provide detailed descriptions by extracting the results at $t = 0.25$, $t = 0.50$, and $t = 0.75$. It can be demonstrated that the predicted solutions are in good agreement with the actual solutions. By using limited unlabeled data, the AutoKE framework can capture the intricate variations of the KdV equation and provide an accurate solution in an automatic manner.

### B. Viscous gravity current

Viscous gravity current is known for describing the long-term behaviors of current front height, and it is often regarded as an asymptotic solution for more complicated physical processes [38], [39]. In this example, our purpose is to test the framework's ability to deal with equations including compound derivatives, i.e., to determine whether the equation parser module can correctly parse and understand complex equation forms, and automatically complete the process of AD.

In this test, we consider the following problem:

$$
\begin{aligned}
&u_t = (uu_x)_x, \quad x \in [1, 2], \ t \in [0, 0.5], \\
&u(x, 0) = -\sin(\pi x), \\
&u(1, t) = u(2, t) = 0,
\end{aligned}
\tag{25}
$$

The ASCII representation of its residual can be represented as "diff(u * diff(u,x), x) - diff(u,t)". To generate the training and test dataset, the finite difference method is utilized to solve the above problem. The number of collocation points required for training is the same as in the previous example. We choose a relatively lightweight neural network structure with 8 hidden layers and 20 nodes per hidden layer since it is not a difficult problem.

It can be observed from Fig. 6 that the AutoKE framework can provide an accurate approximation in the entire data domain with a relative $L_2$ error of 1.02E-2, which means that the whole modeling process is not only automatic, but correct. It is worth noting that practical problems often contain more complex compound items in the equation, which require heavy and time-consuming hand-coded analysis. Our framework, however, can automate the entire modeling process, making it more beneficial for users to solve more complicated problems.

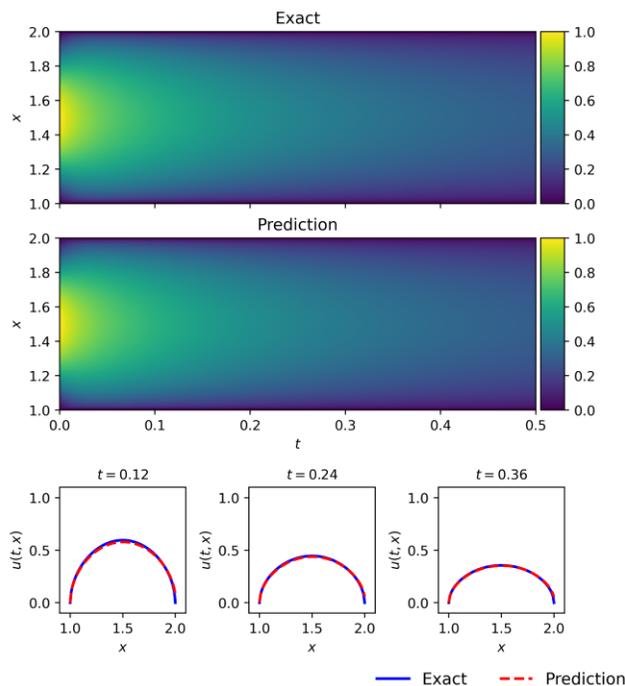

Fig. 6. Numerical approximations of viscous gravity current learned using the AutoKE framework vs. accurate solutions.

*C. Single-phase reservoir flow*

In this experiment, we conduct a specific study on the subsurface single-phase reservoir flow problem that conforms to the governing equation (26), in order to verify the accuracy and efficiency of the framework in solving complex engineering problems. Specifically, compared with the first two one-dimensional problems, this work considers a two-dimensional transient saturated flow, and the nonlinearity of the equation is stronger, which is mainly reflected in: (1) the phase in this problem refers specifically to the oil phase, and its physical parameters vary with the main variable owing to fluid compressibility; and (2) there is a nonlinear and stochastic hydraulic field, and an auxiliary DNN needs to be constructed to approximate the unknown parameters.

$$\frac{\partial}{\partial x}(\frac{K}{B\mu}\frac{\partial P}{\partial x}) + \frac{\partial}{\partial y}(\frac{K}{B\mu}\frac{\partial P}{\partial y}) = \frac{\phi C}{B}\frac{\partial P}{\partial t} \tag{26}$$

where $K$ and $\phi$ denote the permeability and porosity of the reservoir, respectively; $B$ denotes the formation volume factor; and $C$, $P$, and $\mu$ denote the compressibility, pressure, and viscosity of the oil phase, respectively.

In this problem, the entire square domain is evenly divided into 51×51 grids with 1020 m in each direction. The left and right boundaries of the domain are preset fixed pressure boundaries. The left is the inlet with a fixed pressure measured at 300 bar, and the right boundary is the fluid outlet with a fixed pressure measured at 200 bar. The initial formation pressure of the rest of the domain is 200 bar. The upper and lower boundaries are no-flow boundaries. The entire reservoir is isotropic with a fixed porosity of 0.2. It is worth mentioning that the formation volume factor and viscosity of oil are the function of pressure, which satisfy the following physics law:

$$B_P = B_{P_{ref}} \exp[-C(P - P_{ref})] \tag{27}$$

$$\mu_P = \mu_{P_{ref}} \exp[C_v(P - P_{ref})] \tag{28}$$

where $C$ denotes the compressibility of the oil, which takes the value of $9\times10^{-3}$ 1/bar ; $C_v$ denotes the oil viscosibility, which is set to $1\times10^{-2}$ 1/bar ; and $P_{ref}$ denotes the reference pressure, which is set to 200 bar. The formation volume factor and viscosity of the oil at reference pressure are set $1.1\,rm^3/sm^3$ and 2.0 cp, respectively. It is worth noting that introducing these state-dependent parameters into the problem not only increases the difficulty of parsing the equation and implementing physical constraints, but also contributes to stronger nonlinearity. The compact and simplified ASCII representation of the governing equation residual can be expressed as "diff(f * K * diff(P,x),x) + diff(f * K * diff(P,y),y) – diff(P,t)", where f denotes the complex constitutional relationship. Its corresponding AST representation can be found in the Supplementary Materials.

The reference permeability field in this problem is assumed to be a Gaussian random field (GRF), which can be generated by employing Karhunen-Loeve expansion (KLE) [40]. The exponential covariance of the function of $\ln K$ for the two-dimensional stochastic process is shown as follows:

$$C_{\ln K}(x_1, x_2) = \sigma_{\ln K}^2 \exp[-(\frac{|x_1 - x_2|}{\eta_x} + \frac{|y_1 - y_2|}{\eta_y})] \tag{29}$$

where $(x_i, y_i)$ represents the coordinates of the grid block; $\sigma_{\ln K}$ is the standard deviation, which takes the value of 2.0; and $\eta_x$ and $\eta_y$ denotes the correlation length along the x and y direction, respectively, which is set as $\eta_x = \eta_y = 408$. In addition, 32 truncated terms are utilized in KLE to parameterize the random field, which enables 85% of the energy to be preserved. The mean of the field obeys $\langle \ln K \rangle = 4.0\ln(md)$. The concrete details of KLE can be found in Zhang *et al.* [40], [41]. The generated permeability field in this problem is shown in Fig. 7.

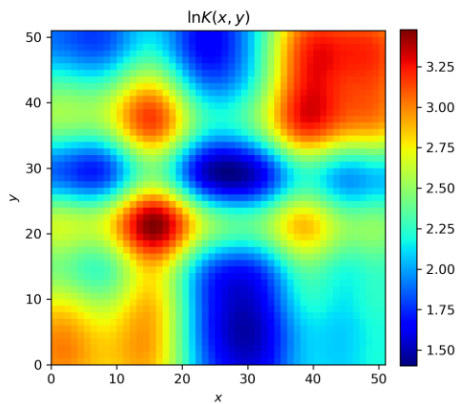

Fig. 7. The log-transformed permeability field of this experiment.

In this case, the reference value of the pressure distribution for the first 50 time steps is calculated by the finite difference method with the time interval $\Delta t = 1$ day. To solve this problem, the neural network adopted in the AutoKE framework consists of 8 hidden layers with 64 neurons in each layer, and the default activation function Softplus is used according to the setup in Wang *et al*. [42]. By default, 1,000 collocation points are randomly extracted from the whole simulation domain to impose equation constraints, 10,000 data points are sampled to meet the initial and boundary conditions, and all of the training datasets are utilized by a mini-batch strategy with a batch size equal to 180. It is worth noting that during the training process, only the Adam optimizer with a learning rate of 0.001 is used to optimize the loss function to ensure fairness of the following ablation study. The training of the network is terminated at 2,000 epochs unless otherwise stated.

### 1) Test of prediction accuracy

In this case, the neural network is trained in a label-free manner, and the pressure distribution of the whole 50 time steps is expected to be extrapolated by the emulator. The weights of each item in the loss function are manually set to 1, without meticulous tuning. As shown in Fig. 8, the prediction values at the last time step are in good agreement with the reference value with an acceptable error. We also provide the comparison of the pressure distribution along the horizontal lines ( $y = 320\ m$, $620\ m$, $920\ m$ ) between the predictions and reference values. The relative $L_2$ error is 3.83E-2, and the coefficient of determination $R^2$ is 0.996. It can be seen that the results are acceptable but not sufficiently accurate, and we will later discuss some improvement measures, such as searching for a more powerful and suitable neural network or incorporating transfer learning.

### 2) Test of prediction accuracy in the presence of noise

In this section, we continue to discuss the robustness of predictions obtained by the framework for solving single-phase reservoir flow problems. We redefine the problem as, given a certain amount of observation data from the previous 18 time steps, predicting the pressure distribution over the next 32 time steps. Consequently, noise can be added to the observation data, which is more in accordance with the general situation in engineering. The specific way to add noise is shown below:

$$P(t, x, y) = P(t, x, y) + P_{diff}(x, y) \times a\% \times \varepsilon \tag{30}$$

where $P_{diff}$ denotes the difference between the maximum value and the minimum value of pressure; $a$ denotes the noise level; and $\varepsilon$ represents a random value sampled from -1 to 1. For each noise level, 10 parallel experiments with different network initializations were performed to increase the reliability of the results.

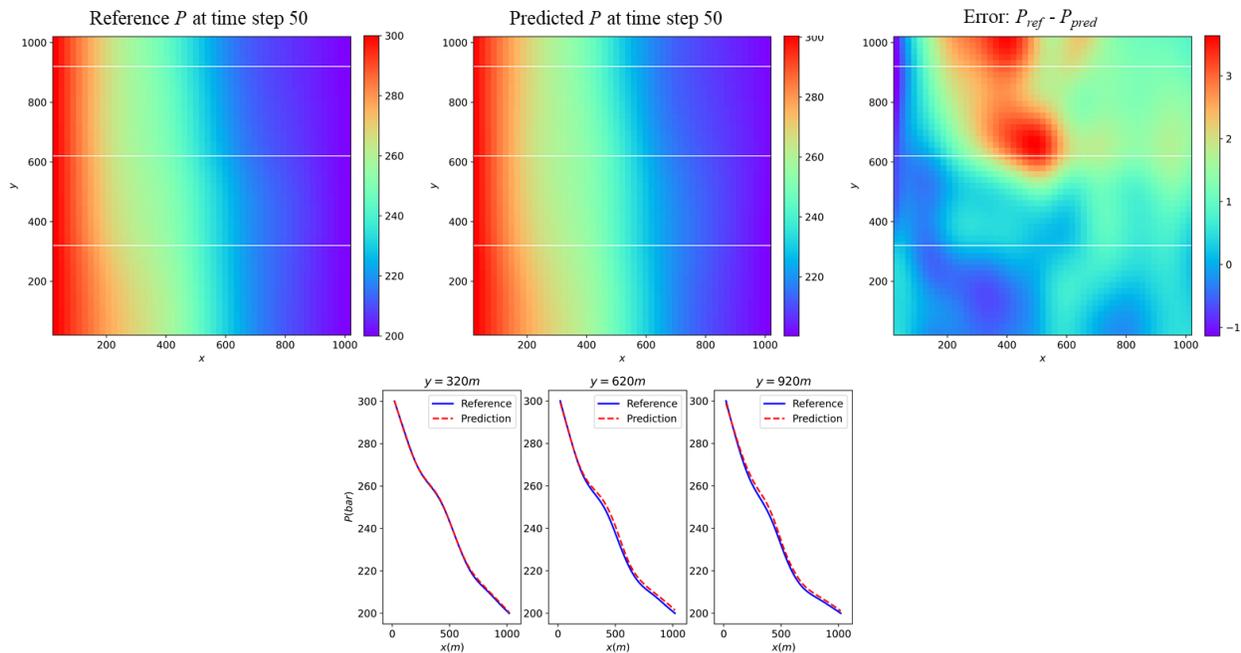

Fig. 8. The pressure prediction at time step 50 using the AutoKE framework and accurate FD method. The upper panel gives the pressure distribution of reference (left), prediction (middle), and the difference between them (right). The lower panel illustrates the pressure along y = 320 m, 620 m, and 920 m. The red dashed curves represent the prediction values, and the blue solid curves refer to the reference values.

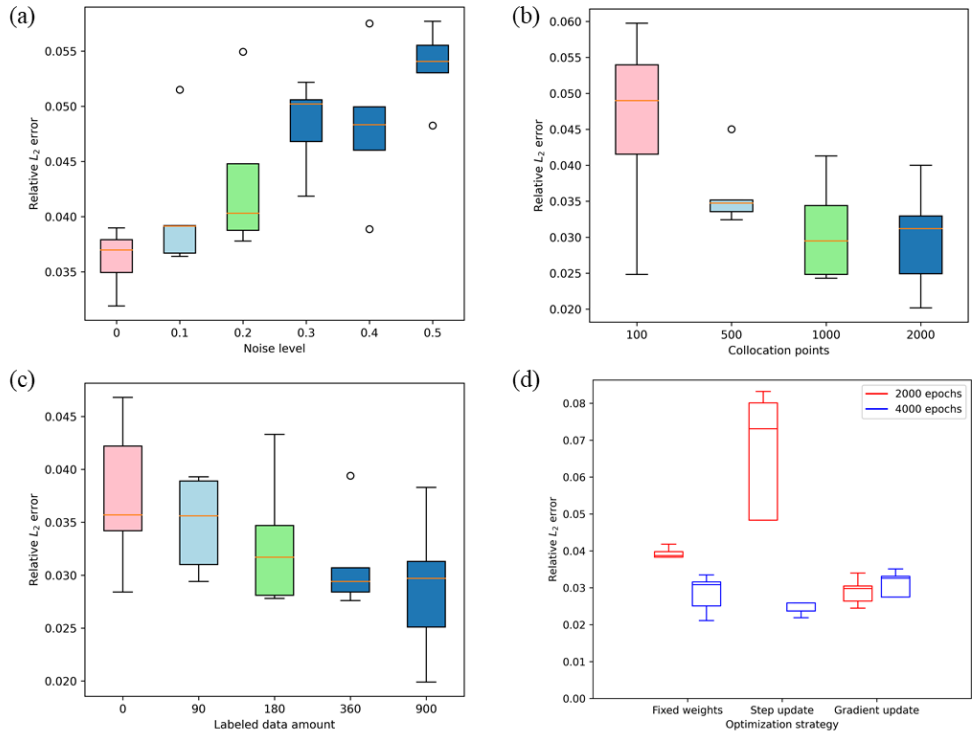

Fig. 9. Boxplot of relative L$_2$ error of predictions with (a) different noise; (b) different numbers of collocation points; (c) different numbers of observation data; and (d) different optimization parameter adjustment strategies.

Fig. 9a illustrates the performance of the framework under different noise levels. It is found that as the noise level continues to increase, the accuracy of the prediction will decrease but remains within an acceptable range. It can be inferred from the result that the embedding of physical knowledge greatly increases the model's ability to resist noise disturbance and provides a more robust prediction.

### 3) Test of prediction accuracy with different numbers of training data

Two types of data, including collocation points and observation data, are considered here in order to test the influence of training data on the emulator's extrapolation performance. We kept the number of collocation points to be 100, 500, 1,000, and 2,000, respectively, for each training process, all of which are randomly sampled inside the simulation domain.

It can be seen from Fig. 9b that increasing the number of collocation points will dramatically improve the accuracy of the predictions at the beginning, but when the number of points reaches 1,000, the effect of improvement is no longer obvious. In addition, we also test the sensitivity of the model performance to the observation data. Since incorporating the observation data transforms the original problem into predicting the future response based on the data from the first 18 time steps, it would be much easier for the model to learn the physical knowledge. Fig. 9c shows that the relative L$_2$ error decreases as the observation data size increases. It is worth noting that increasing the training data can improve the accuracy of prediction, but it will also increase the training time. For specific instances, a balance needs to be found between them.

### 4) Performance of adaptive parameters

In this part, we intend to exploit the Lagrangian dual method to help solve the constraint optimization problems. Since only the errors on unknown permeability field relations and governing equation matter to the final prediction, we set adaptive multipliers $\lambda_k$ and $\lambda_r$ with an initial value of 1, and compared this method with common soft constraint implementations with fixed coefficients determined by experience. We set the step size $s_k$ and $s_r$ to be 1.1 and 1.25, respectively, for the step update method. The learning rate for the gradient ascent strategy is 0.001. It is worth noting that as these adaptive optimization parameters are utilized to reduce the repetitive experiments, all of these hyperparameters are randomly selected with a small value. The training time is also extended to 4,000 epochs for a more detailed analysis. Five parallel experiments for each setting are conducted with different neural network initializations and sampled training data. It can be seen from Fig. 9d that using the adaptive parameter adjustment strategy based on gradient ascent can make the model converge faster with a more accurate prediction result compared with common fixed parameters. The step update strategy with a casually selected step size cannot converge rapidly, but still obtains a good prediction after extending the training epoch. Actually, the adaptive parameter adjustment method proposed in the AutoKE framework does not necessarily acquire better prediction results than sophisticated manual parameter tuning. The purpose is to lower the threshold for application and conserve the time spent on repeated parameter tuning experiments. Indeed, it can be seen that users can often obtain satisfactory results by adopting a more conservative update strategy with only one attempt.

### 5) *Performance of NAS*

The former experiments show that the framework can provide relatively accurate results based on the default and randomly selected architecture and hyperparameters. In this part, a test on searching for a more suitable structure of the network is investigated to further improve the prediction performance. In fact, only three hyperparameters need to be considered in this case, including layer number of feedforward neural network (FFN), activation function, and hidden unit size. The predefined parameters set are listed in Table I, and the options in bold black are the default parameter settings.

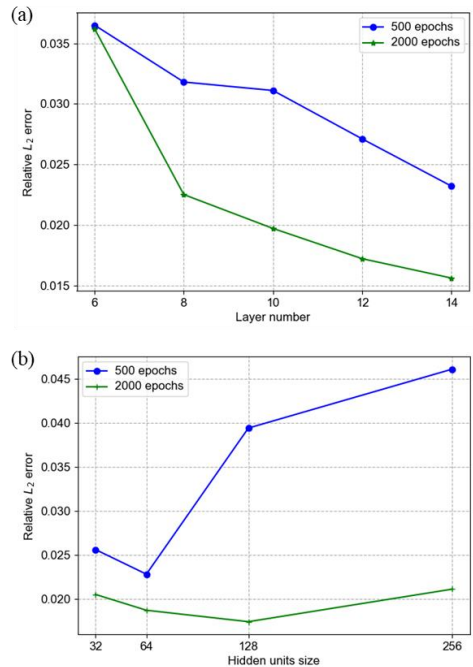

Fig. 10. Relative $L_2$ error with different (a) FFN layer number; (b) and hidden unit size.

To select a better neural network architecture, the grid search algorithm is utilized in the AutoKE framework, which is a naïve method based on greedy search. Two criteria can be used to evaluate the model's performance. One is to randomly select some labeled data, such as reference values of the first 18 time steps, as the development set and the other one is the residual of the governing equation. In this case, the former is adopted,

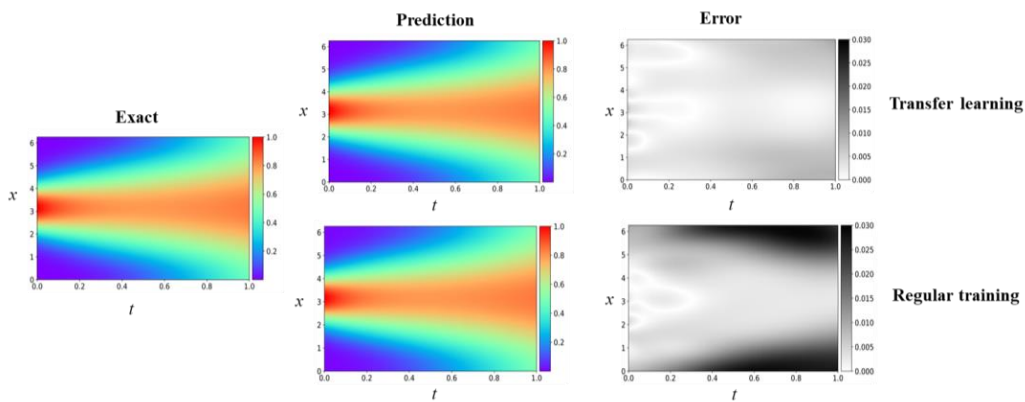

Fig. 11. Prediction results of transfer learning and regular training for $\nu = 1$, and $\rho = 2$.

and the early stopping strategy is used to reduce the training time.

It can be seen from Fig. 10 and Table II that default parameters are not the optimal combination, and a more suitable neural architecture can be used for this case. For instance, increasing the layer number and hidden unit dimension can both improve the model performance, but also require a larger training time. It is worth noting that the performance of the emulator at 500 and 2,000 epochs is inconsistent, since it tends to take more time to converge for models with more parameters. As a consequence, the early stopping strategy needs to be used with caution. In addition, GeLU is a more suitable choice for the activation function according to Table II. In the actual training process, we need to find a balance between model size and prediction accuracy, and the AutoKE framework is able to automate the searching process and facilitate the whole training process.

TABLE I
HYPERPARAMETERS SET FOR NAS

| Hyperparameters | Predefined parameters set |
|---|---|
| FFN layer number | {6, **8**, 10, 12, 14} |
| Activation function | {**SoftPlus**, Tanh, ReLU, GELU} |
| Hidden unit size | {32, **64**, 128,256} |

TABLE II
PREDICTION PERFORMANCE WITH DIFFERENT ACTIVATION FUNCTIONS

| Activation function | Relative $L_2$ error | $R^2$ score |
|---|---|---|
| Softplus | $3.83 \times 10^{-2}$ | $9.967 \times 10^{-1}$ |
| Tanh | $4.30 \times 10^{-2}$ | $9.961 \times 10^{-1}$ |
| ReLU | $4.25 \times 10^{-2}$ | $9.961 \times 10^{-1}$ |
| GeLU | $\mathbf{3.18 \times 10^{-2}}$ | $\mathbf{9.979 \times 10^{-1}}$ |

## D. Transfer learning

One important feature of the AutoKE framework is that it can extend the trained model or emulator to solve similar problems. In this part, we intend to test the function and role of transfer learning in accelerating training and even solving more complex physical problems.

### 1) Test of the reaction-diffusion equation

A well-trained neural network is obtained in the second example of viscous gravity current. Now, we consider a new system, named reaction-diffusion:

$$u_t = vu_{xx} + \rho u(1-u), x \in \Omega, \ t \in (0,T]$$
$$u(x,0) = h(x), \quad x \in \Omega$$

(31)

where $v$ is the diffusion coefficient; and $\rho$ denotes the reaction coefficient. In this case, $v$ is set to be 1, and $\rho$ is variable. This equation can be solved analytically to obtain the final solution. For performing transfer learning, the whole weights and biases of the neural network are inherited from the viscous gravity current case, and all of the parameters are trainable in the training process.

Table III shows that transfer learning can significantly reduce the absolute/relative error compared with regular training which is trained from scratch when the reaction coefficient is not sufficiently large. The diffusion term in the equation plays a more significant role in the equation when $\rho$ is larger, which contributes to a more substantial difference in data distribution between the viscous gravity current case and the current problem. It is obvious that the network is stuck at a local minimum by transfer learning and performs poorly when $\rho$ is equal to 3. When the problem to be solved is relatively similar to the original problem with a well-trained emulator, transfer learning can provide a more accurate prediction result, as shown in Fig. 11, and dramatically increase the convergence rate, as shown in Fig. 12.

TABLE III
PREDICTIONS OF REGULAR TRAINING AND TRANSFER LEARNING FOR DIFFERENT COEFFICIENTS

| | | Regular training | Transfer learning |
|---|---|---|---|
| $v = 1, \rho = 0.5$ | absolute error | $2.35 \times 10^{-3}$ | $\mathbf{1.30 \times 10^{-3}}$ |
| | $L_2$ relative error | $7.63 \times 10^{-3}$ | $\mathbf{3.69 \times 10^{-3}}$ |
| $v = 1, \rho = 1$ | absolute error | $4.15 \times 10^{-3}$ | $\mathbf{1.84 \times 10^{-3}}$ |
| | $L_2$ relative error | $1.17 \times 10^{-2}$ | $\mathbf{4.82 \times 10^{-3}}$ |
| $v = 1, \rho = 2$ | absolute error | $9.38 \times 10^{-3}$ | $\mathbf{3.23 \times 10^{-3}}$ |
| | $L_2$ relative error | $2.33 \times 10^{-2}$ | $\mathbf{7.25 \times 10^{-3}}$ |
| $v = 1, \rho = 3$ | absolute error | $\mathbf{1.46 \times 10^{-2}}$ | $2.05 \times 10^{-1}$ |
| | $L_2$ relative error | $\mathbf{3.26 \times 10^{-2}}$ | $5.03 \times 10^{-1}$ |

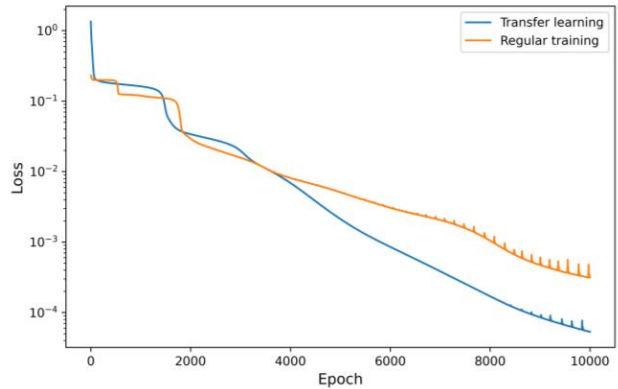

Fig. 12. Loss of transfer learning and regular training.

### 2) Test of single-phase reservoir flow with curriculum learning

For solving the single-phase reservoir flow problem, regular training can obtain a relatively good, but not sufficiently accurate, prediction. According to the results of NAS, an increment of network complexity, accomplished by increasing FFN layer numbers and hidden unit dimension, can improve the prediction performance, but at the same time lead to a longer training time. In this case, we test whether transfer learning can achieve more accurate predictions in less time. The base task adopted to generate the pre-trained model is the common groundwater flow problem, which has been well researched and discussed in many references [42]-[44]. Compared with single-phase flow in oil reservoirs, it is much simpler without strong nonlinearity. We keep the same network structure for a fair comparison. A detailed description of the prediction results can be found in the Supplementary Materials.

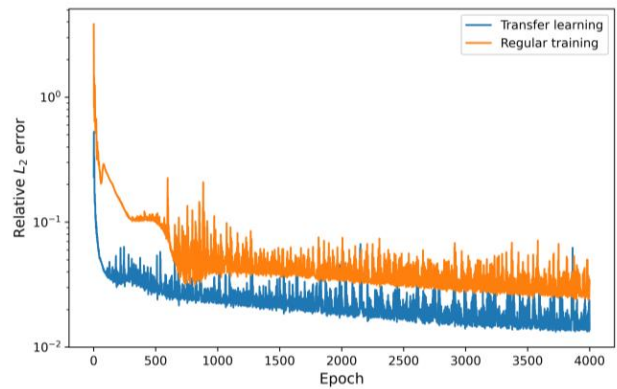

Fig. 13. The relative $L_2$ error of transfer learning and regular training.

Fig. 13 shows that performing transfer learning accelerates the training process, and it always has a smaller relative $L_2$ error during the whole training process. Table IV provides more straightforward comparisons, and the network initialized from a different subsurface flow problem can make a more accurate prediction after only 500 epochs of training than the network with random initialization after 2,000 epochs. According to this example, one can solve complicated physical problems by starting from a similar but simple problem, which is identical to the curriculum learning. Then, the neural network can be initialized with the well-trained one, and it can make more accurate predictions with much less training time.

TABLE IV
THE RELATIVE $L_2$ ERROR OF REGULAR TRAINING AND TRANSFER LEARNING

| Epoch | Regular training | Transfer learning |
|-------|------------------|-------------------|
| 500   | $5.21 \times 10^{-2}$ | $\mathbf{2.86 \times 10^{-2}}$ |
| 1000  | $4.10 \times 10^{-2}$ | $\mathbf{2.63 \times 10^{-2}}$ |
| 1500  | $3.88 \times 10^{-2}$ | $\mathbf{1.90 \times 10^{-2}}$ |
| 2000  | $3.83 \times 10^{-2}$ | $\mathbf{1.38 \times 10^{-2}}$ |

## V. CONCLUSIONS AND FUTURE WORK

This paper proposes an efficient and automatic machine learning framework (AutoKE) for embedding knowledge into the neural network. Taking the subsurface reservoir flow problem as an example, we first analyze the problems and inconveniences of embedding physical knowledge and then introduce the advantages of this framework, which can be summarized as follows:

- Automate the process of embedding physical knowledge. Compared with the existing methods, this framework approximates unknown physics in the equation by introducing a series of fitting networks that can be jointly trained with the prediction network. In addition, we use the equation parser module to parse the ASCII representation of the input equation and convert it to a binary abstract syntax tree. The construction of the computational graph is accomplished by traversing the nodes of the tree, and then the residual of the equation can be automatically calculated by AD.

- In the design of neural networks, we implement NAS through a simple grid search algorithm to help users find suitable DNNs for solving the current physical problem.
- The Lagrangian dual method is used to replace the fixed weights in the original loss equation with adaptive optimization parameters, which can be automatically adjusted during the training process according to the corresponding loss value, avoiding the complicated and cumbersome parameter tuning process.
- We further explore the application of transfer learning in the scalability of the trained emulator, which facilitates the training process and contributes to better prediction results for solving similar problems. In addition, from the perspective of curriculum learning, a new method is proposed to solve complex physical problems through a process from simple to difficult.

In the experiment part, a series of physical problems, including the KdV equation, viscous gravity current, and single-phase reservoir flow, is tested by using the AutoKE framework. The results show that this framework can not only realize automatic establishment of the emulator to solve equations, but also achieve high prediction accuracy and excellent training efficiency.

It is worth noting that the functions of the framework proposed in this paper and other PINN toolboxes, such as DeepXDE, do not overlap, but rather possess a good complementary relationship. In future work, we will further improve this framework, and provide more complete solutions and program interfaces for different boundary conditions and physical constraints. Moreover, it can be seen that when searching for the appropriate framework, we use a simple grid search method. Although the early stopping strategy can save computational expense to a certain extent, the termination time is difficult to determine. In the future, we intend to explore more efficient and intelligent searching methods, such as gradient-based method DARTs [45].

**Code and data availability**
The code and data of this article can be found at https://github.com/menggedu/AutoKE.

ACKNOWLEDGMENT

The authors express gratitude to Mr. Nanzhe Wang and Mr. Jian Li for their assistance with data preparation, and constructive suggestions during the course of this article.

# Supplementary Materials for

# AutoKE: An automatic knowledge embedding framework for scientific machine learning

S1 Intermediate results

Here, we provide the intermediate results of each case in the numerical experiments. When using the AutoKE framework, the user first enters an ASCII representation of the equation and it can be converted into an abstract syntax tree (AST), which is a structural representation containing all of the information. The AST of the Korteweg-de Vries equation (KdV) and viscous gravity current are shown below in Fig. S1.

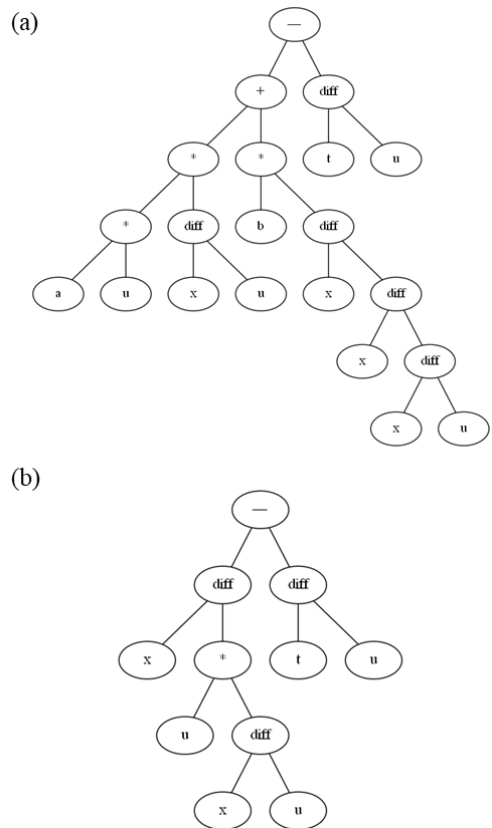

Fig. S1. ASTs created by the equation parser for (a) KdV equation; and (b) viscous gravity current.

For the single-phase reservoir flow problem, it is much more complex since there is a known state-dependent constitutional relationship $f$ and a space-dependent diffusion coefficient $K$ in the compound derivatives. Its compact AST representation is shown in Fig. S2.

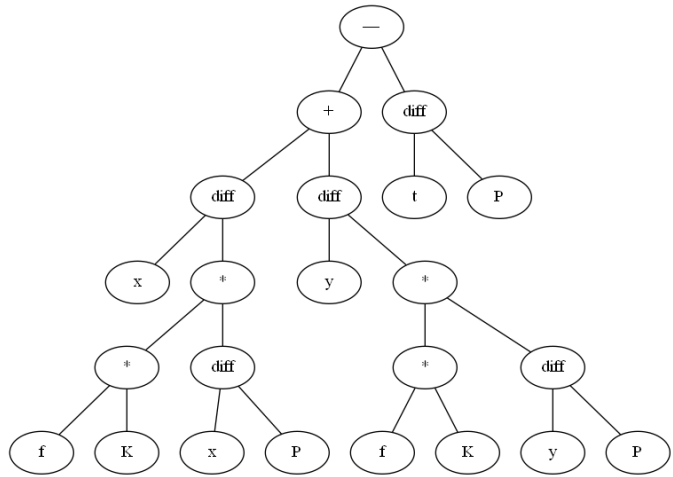

Fig. S2. The AST created by the equation parser for single-phase reservoir flow.

## S2 Groundwater flow problem for transfer learning

We take a common 2D unsteady-state groundwater flow problem as the base task to investigate the effect of transfer learning. The governing equation can be expressed as follows:

$$S_s \frac{\partial h}{\partial t} = \frac{\partial}{\partial x}(K(x, y)\frac{\partial h}{\partial x}) + \frac{\partial}{\partial y}(K(x, y)\frac{\partial h}{\partial y}) \tag{32}$$

where $h$ refers to the hydraulic head, which is the state of interest; $S_s$ denotes the specific storage and $S_s = 0.0001$ [$L^{-1}$], where L refers to the consistent length unit; and $K(x, y)$ represents the hydraulic conductivity. Karhunen-Loeve expansion is utilized to generate the random field $K$ again, but with different mean and variance values [1], [2]. The log hydraulic conductivity satisfies the distribution with $\langle \ln K \rangle = 0$, $\sigma_{\ln K}^2 = 1$. The geometry of the simulation domain is the same as that in the target task, but the specific boundary conditions and initial conditions are different. The prescribed hydraulic heads on the left and right boundaries are $H_{x=0} = 202[L]$ and $H_{x=1020} = 200[L]$, respectively. At the initial condition, the hydraulic head takes values of $H_{t=0,x=0} = 202[L]$ and $H_{t=0,x\neq0} = 200[L]$, respectively. Note that the setup of this problem is exactly the same as that in Wang *et al*. [3], where more detailed descriptions of this problem can be found.

The results of hydraulic head at time step 50 are shown in Fig. S3. It can be observed that the AutoKE framework can make an accurate prediction for this relatively simple problem. Based on this task, we utilize the transferred features to initialize the neural networks used for solving single-phase flow in oil reservoirs. The results show that making use of the output obtained from a well-solved simple problem is of great benefit to solving more complicated problems.

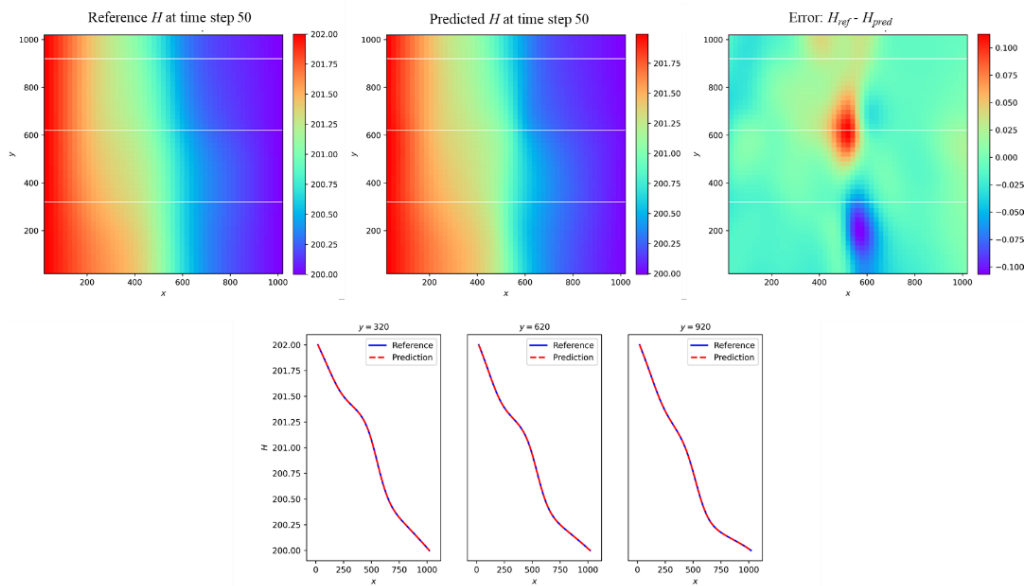

Fig. S3. Prediction results of the base task at time step 50